\pdfoutput=1
\documentclass[11pt]{article}

\usepackage[]{EMNLP2022}

\usepackage{times}
\usepackage{latexsym}

\usepackage[T1]{fontenc}
\usepackage[utf8]{inputenc}

\usepackage{microtype}
\usepackage{algorithm}
\usepackage{algorithmic}
\usepackage{booktabs}
\usepackage{amsmath}
\usepackage{multirow}
\usepackage{bbding}
\usepackage{mathrsfs}
\usepackage{subfigure}
\usepackage{amssymb}
\usepackage{makecell}
\usepackage{graphicx} % DO NOT CHANGE THIS

\usepackage{inconsolata}

\title{UniNL: Aligning Representation Learning with Scoring Function for OOD Detection via Unified Neighborhood Learning}

\author{Yutao Mou$^{1*}$, Pei Wang$^{1*}$, Keqing He$^{2*}$, Yanan Wu$^{1}$ \\
{\bf Jingang Wang$^{2}$,} {\bf Wei Wu$^{2}$,} {\bf Weiran Xu$^{1}$}\thanks{\ \ The first three authors contribute equally. Weiran Xu is the corresponding author.}\\
  $^1$Beijing University of Posts and Telecommunications, Beijing, China\\
$^{2}$Meituan, Beijing, China\\
  \texttt{\{myt,wangpei,yanan.wu,xuweiran\}@bupt.edu.cn}\\
  \texttt{\{hekeqing,wangjingang,wuwei\}@meituan.com}
}

\begin{document}
\maketitle
\begin{abstract}
Detecting out-of-domain (OOD) intents from user queries is essential for avoiding wrong operations in task-oriented dialogue systems. The key challenge is how to distinguish in-domain (IND) and OOD intents. Previous methods ignore the alignment between representation learning and scoring function, limiting the OOD detection performance. In this paper, we propose a unified neighborhood learning framework (UniNL) to detect OOD intents. Specifically, we design a K-nearest neighbor contrastive learning (KNCL) objective for representation learning and introduce a KNN-based scoring function for OOD detection. We aim to align representation learning with scoring function. Experiments and analysis on two benchmark datasets show the effectiveness of our method. \footnote{We release our code at \url{https://github.com/Yupei-Wang/UniNL}}

\end{abstract}

\section{Introduction}

Out-of-domain (OOD) intent detection aims to know when a user query falls outside the range of pre-defined supported intents, which helps to avoid performing wrong operations and provide potential directions of future development in a task-oriented dialogue system \cite{Akasaki2017ChatDI,Tulshan2018SurveyOV,Shum2018FromET,Lin2019DeepUI,xu-etal-2020-deep,zeng-etal-2021-modeling,zeng-etal-2021-adversarial}. Compared with normal intent detection tasks, we don't know the exact number and lack labeled data for unknown intents, which makes it challenging to identify OOD samples in the task-oriented dialog.

Previous OOD detection works can be generally classified into two types: supervised \cite{Fei2016BreakingTC,Kim2018JointLO,Larson2019AnED,Zheng2020OutofDomainDF} and unsupervised \cite{Bendale2016TowardsOS,Hendrycks2017ABF,Shu2017DOCDO,Lee2018ASU,Ren2019LikelihoodRF,Lin2019DeepUI,xu-etal-2020-deep,zeng-etal-2021-modeling} OOD detection. The former indicates that there are extensive labeled OOD samples in the training data. \newcite{Fei2016BreakingTC,Larson2019AnED}, form a \emph{(N+1)}-class classification problem where the \emph{(N+1)}-th class represents the OOD intents. Further, \newcite{Zheng2020OutofDomainDF} uses labeled OOD data to generate an entropy regularization term. But these methods require numerous labeled OOD intents to get superior performance, which is unrealistic. We focus on the unsupervised OOD detection setting where labeled OOD samples are not available for training. Unsupervised OOD detection first learns discriminative representations only using labeled IND data and then employs scoring functions, such as Maximum Softmax Probability (MSP) \cite{Hendrycks2017ABF}, Local Outlier Factor (LOF) \cite{Lin2019DeepUI}, Gaussian Discriminant Analysis (GDA) \cite{xu-etal-2020-deep} to estimate the confidence score of a test query. 

All these unsupervised OOD detection methods only focus on the improvement of a single aspect of representation learning or scoring function, but none of them consider how to align representation learning with scoring functions. For example, \newcite{Lin2019DeepUI} proposes a local outlier factor for OOD detection, which considers the local density of a test query to determine whether it belongs to an OOD intent, but the IND pre-training objective LMCL \cite{wang2018cosface} cannot learn neighborhood discriminative representations.  \newcite{xu-etal-2020-deep,zeng-etal-2021-modeling} employ a gaussian discriminant analysis method for OOD detection, which assumes that the IND cluster distribution is a gaussian distribution, but they use a cross-entropy or supervised contrastive learning \cite{Khosla2020SupervisedCL} objective for representation learning, which cannot guarantee that such an assumption is satisfied. The gap between representation learning and scoring function limits the overall performance of these methods.

To solve the conflict, in this paper, we propose a \textbf{Uni}fied \textbf{N}eighborhood \textbf{L}earning framework (\textbf{UniNL}) for OOD detection, which aims to align IND pre-training representation objectives with OOD scoring functions. Our intuition is to learn neighborhood knowledge \cite{breunig2000lof} to detect OOD intents. For IND pre-training, we introduce a K-Nearest Neighbor Contrastive Learning Objective (KNCL) to learn neighborhood discriminative representations. Compared to SCL \cite{zeng-etal-2021-modeling} which draws all samples of the same class closer, KNCL only pulls together similar samples in the neighbors. To align KNCL, we further propose a K-nearest neighbor scoring function, which estimates the test sample confidence score by computing the average distance between a test sample and its K-nearest neighbor samples. The KNCL objective learns neighborhood discriminative knowledge, which is more beneficial to promoting KNN-based scoring functions.

Our contributions are three-fold: (1) We propose a unified neighborhood learning framework (UniNL) for OOD detection, which aims to match IND pre-training objectives with OOD scoring functions. (2) We propose a K-nearest neighbor contrastive learning (KNCL) objective for IND pre-training to learn neighborhood discriminative knowledge, and a KNN-based scoring function to detect OOD intents. (3) Experiments and analysis demonstrate the effectiveness of our method for OOD detection.

\section{Approach}
\label{method}

\begin{figure}[t]
    \centering
    \resizebox{.5\textwidth}{!}{
    \includegraphics{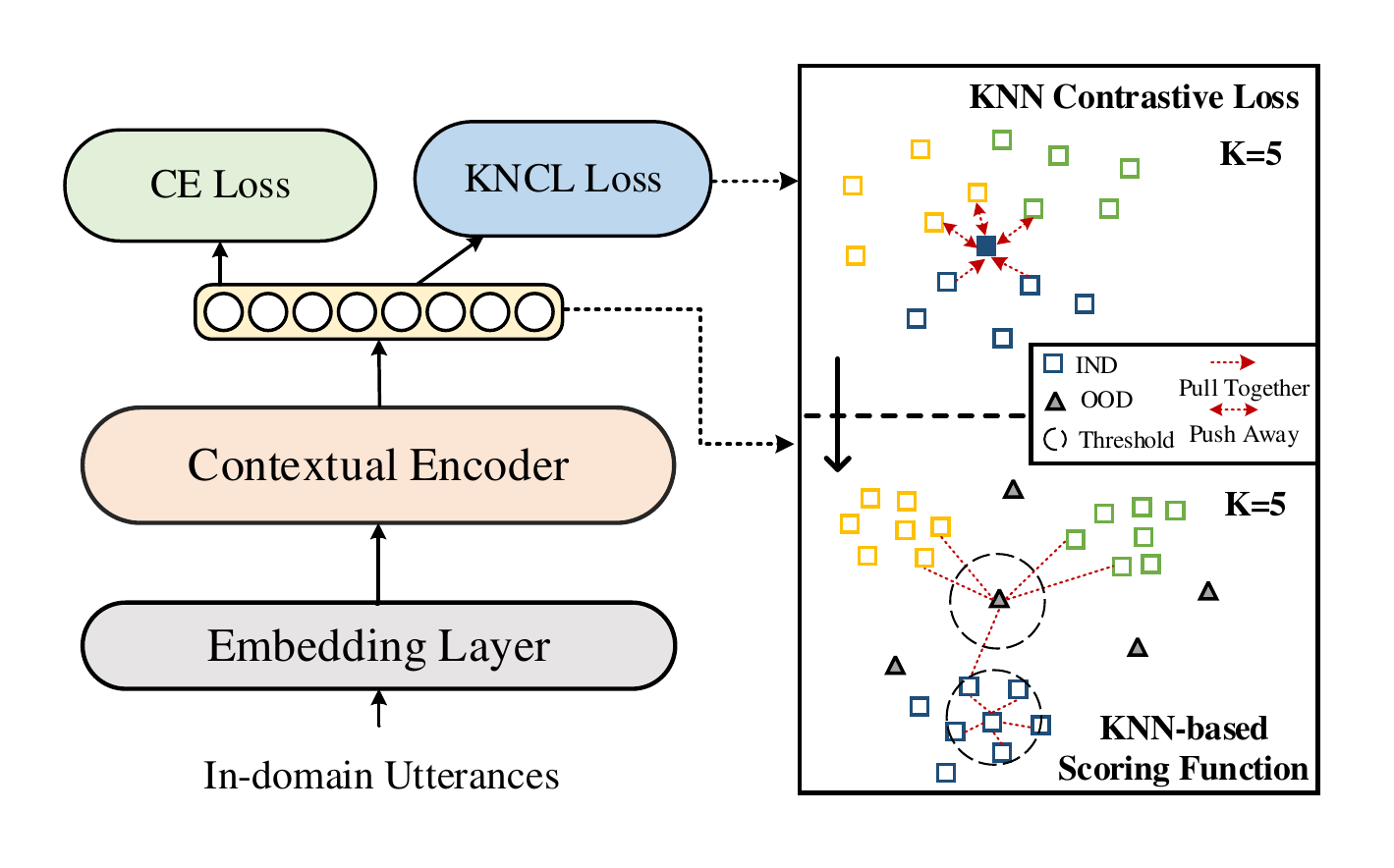}}
    %\vspace{-0.9cm}
    \caption{Overall architecture of UniNL.}
    \label{fig:model}
    \vspace{-0.5cm}
\end{figure}

\textbf{Overall Architecture} Fig \ref{fig:model} shows the overall architecture of our proposed UniNL, which includes K-nearest contrastive learning (KNCL) and KNN-based score function. We first train an in-domain intent classifier using our KNCL objective in the training stage, which aims to learn neighborhood discriminative representation. Then in the test stage, we extract the intent feature of a test query and employ our proposed KNN-based score function to estimate the confidence score. We aim to align representation learning and scoring functions. 

\begin{table*}[t]
\centering
\resizebox{0.80\textwidth}{!}{%
    \begin{tabular}{c|c|cc|cc|cc|cc}
    % \hline
    % \toprule[1pt]
    \hline
    \multirow{3}{*}{Detection} & \multirow{3}{*}{Training} & \multicolumn{4}{c|}{CLINC-Full}& \multicolumn{4}{|c}{CLINC-Small} \\\cline{3-10}
        ~ & ~ & \multicolumn{2}{c|}{IND}  & \multicolumn{2}{c|}{OOD} & \multicolumn{2}{c|}{IND}  & \multicolumn{2}{|c}{OOD} \\  \cline{3-10}
        ~ & ~ & \multicolumn{1}{c}{ACC} & \multicolumn{1}{c|}{F1} & \multicolumn{1}{c}{Recall}& \multicolumn{1}{c|}{F1} & \multicolumn{1}{c}{ACC}  & \multicolumn{1}{c|}{F1} & \multicolumn{1}{c}{Recall} & \multicolumn{1}{c}{F1}\\ \hline
        \multirow{3}{*}{MSP} & CE & 91.21	&87.75	&45.28	&54.90  &88.98	&86.22	&45.16	&52.04  \\ 
        ~ & SCL &91.10	&87.69	&47.98	&59.64   &89.62	&85.04	&34.64	&47.33 \\ 
        ~ & KNCL(ours) &91.09	&87.05	&\textbf{58.73}	&\textbf{66.98}   &90.15	&86.63	&\textbf{47.26}	&\textbf{59.55} \\ \hline
        \multirow{3}{*}{LOF} & CE &85.46	&85.80	&57.40	&58.78   &82.45	&82.73	&52.88	&53.90   \\ 
        ~ & SCL &86.52	&86.80	&60.72	&61.80   &83.13	&83.39	&56.88	&57.48  \\ 
        ~ & KNCL(ours) &85.77	&85.05	&\textbf{70.10}	&\textbf{66.30}   &86.13	&85.6	&\textbf{60.92}	&\textbf{62.63}  \\ \hline
        \multirow{3}{*}{GDA} & CE &86.34	&87.73	&63.72	&65.23   &84.24	&84.30	&60.40	&61.07   \\ 
        ~ & SCL &\underline{87.01}	&\underline{88.28}	&\underline{66.80}	&\underline{67.68}   &\underline{87.07}	&\underline{86.54}	&\underline{61.46}	&\underline{63.83}  \\ 
        ~ & KNCL(ours) &88.45	&87.08	&\textbf{71.59}	&\textbf{70.37}   &87.15	&87.53	&\textbf{64.00}	&\textbf{66.18}  \\ \hline
        \multirow{3}{*}{KNN(ours)} & CE   &90.30	&88.89	&67.03	 &72.32  &88.64	&86.85	&60.48	&66.65   \\ 
        ~ & SCL &89.42	&88.69	&71.45	&74.17   &89.45	&87.35	&62.23	&70.42 \\ 
        ~ & KNCL(ours)   &91.24	&88.28	&\textbf{72.08}	&\textbf{76.53}   &89.32	&87.61	&\textbf{66.00}	&\textbf{72.79} \\ \hline
        
    % \bottomrule[1pt]
    \end{tabular}%
}
%\vspace{-0.2cm}
\caption{The performance of different OOD scoring functions and IND pre-training objectives on CLINC-Full and CLINC-Small datasets for the BiLSTM-based model (p<0.01 under t-test). The last line is our full UniNL model.}
%\vspace{-0.45cm}
\label{main}
\end{table*}

\textbf{KNN Contrastive Representation Learning} Existing OOD detection methods generally adopt cross-entropy (CE) \cite{xu-etal-2020-deep} and supervised contrastive learning (SCL) \cite{zeng-etal-2021-modeling} objectives for representation learning. Both CE and SCL tend to bring all samples of the same classes closer, and samples of different classes are pushed away. They learn inter-class discriminative features in a global representation space. However, we find that when performing OOD detection, we care more about the data distribution within the neighborhood of a given sample. Inspired by 
\citet{Breunig2000LOFID}, we hope to learn neighborhood discriminative knowledge in the IND pre-training stage to facilitate OOD detection. We propose a K-nearest neighborhood contrastive learning (KNCL) objective to learn discriminative features in a local representation space. Given an IND sample $x_{i}$, we firstly obtain its intent representation $z_{i}$ using a BiLSTM \cite{Hochreiter1997LongSM} or BERT \cite{Devlin2019BERTPO} encoder. Next, we perform KNCL as follows:
{\setlength{\abovedisplayskip}{0.05cm}
\setlength{\belowdisplayskip}{0.2cm}
\begin{align}
\begin{split}
    \mathcal{L}_{K N
    C L}=&\sum_{i=1}^{N}-\frac{1}{|{N}_{k}(i)|} \sum_{j=1}^{{N}_{k}(i)} \mathbf{1}_{i \neq j} \mathbf{1}_{y_{i}=y_{j}} \\ &\log \frac{\exp \left(z_{i} \cdot z_{j} / \tau\right)}{\sum_{k=1}^{{N}_{k}(i)} \mathbf{1}_{i \neq k} \exp \left(z_{i}\cdot z_{k} / \tau\right)}
\end{split}
\end{align}}
where $N_{k}(i)$ is the KNN set of $z_{i}$ in the representation space. KNCL only draws closer together samples of the same class in the neighborhood. Specifically, given an anchor, KNCL first finds its KNN set in a batch, and then selects samples of the same class as positives, and different classes as negatives. Similar to \citet{zeng-etal-2021-modeling}, we use an adversarial augmentation strategy to generate augmented views of the original samples within a batch. In the implementation, we first pre-train the intent classifier using KNCL, then finetune the model using CE, both on the IND data. We leave the implementation details in the appendix.
Section \ref{Qualitative} proves that KNCL learns neighborhood discriminative knowledge and helps to distinguish IND from OOD.

\textbf{KNN-based Score Function} 
To align with the KNCL representation learning objective, we propose a KNN-based scoring function, which makes full use of the neighborhood data distribution to estimate confidence scores. Specifically, given a test query $x_{i}$, we first obtain its intent representation $z_{i}$ through the pre-trained encoder, and then perform L2 normalization. For each sample in the test set, we find its KNN set from the training set, and then calculate the average Euclidean distance as the scoring function. The formula of the KNN-based scoring function is as follows:
{\setlength{\abovedisplayskip}{0.2cm}
\setlength{\belowdisplayskip}{0.2cm}
\begin{align}
    \mathcal{G}_{\lambda}(x_{i})=& \begin{cases}{IND} & \text { if } {\mathcal{S}(x_{i})} < \lambda \\ {OOD} & \text { if } {\mathcal{S}(x_{i})} \geq \lambda \end{cases} \\
{S}(x_{i}) =&\frac{1}{|{N}_{k}(x_{i})|} \sum_{j=1}^{{N}_{k}(x_{i})}{||z_{i}-z_{j}||_2}
\end{align}}
where $N_{k}(x_i)$ is the KNN set of test query $x_{i}$ from the training set, $\lambda$ is the threshold, and we use the best IND F1 scores on the validation set to calculate the threshold adaptively. The KNN-based scoring function needs to consider the data distribution in the neighborhood of a test query to determine whether it is an OOD sample, and the KNCL objective function distinguishes samples of different classes in the neighborhood, so we believe that KNCL representation learning objective aligns with the KNN-based scoring function, which is beneficial to improve the OOD detection performance. We discuss it in section \ref{Qualitative}.

\section{Experiments}

\subsection{Setup}
\textbf{Datasets} We perform experiments on four public benchmark OOD datasets, CLINC-Full, CLINC-Small \cite{Larson2019AnED}, Banking \cite{casanueva2020efficient} and Stackoverflow \cite{xu2015short}. \textbf{Metrics} We use four common metrics for OOD detection to measure the performance, including IND metrics: Accuracy and macro F1, and OOD metrics: Recall and F1. OOD Recall and F1 are the main evaluation metrics. \textbf{Baselines} We compare UniNL with different pre-training objectives CE and SCL, and scoring functions including MSP, LOF and GDA. Besides, we also compare our model with the following state-of-the-art baselines,  SCL\cite{zeng-etal-2021-modeling}, Energy\cite{ouyang-etal-2021-energy}, ADB\cite{Zhang2021DeepOI} and KNN-CL\cite{zhou-etal-2022-knn}. For a fair comparison, we use the same BiLSTM and BERT as backbone. We provide a more comprehensive comparison and implementation details of these models in Appendix \ref{setups}.

% For a fair comparison, we use BiLSTM as backbone network, and compare the effect of KNCL with SCL and CE pre-training objective. To evaluate the effectiveness of our framework, we comprehensively compare the performance of our proposed KNN-based detection algorithm with MSP, LOF, GDA under three different training objectives. Besides, we compare our models with the following state-of-the-art baselines,  SCL\cite{zeng-etal-2021-modeling}, Energy\cite{ouyang-etal-2021-energy}, ADB\cite{Zhang2021DeepOI} and KNN-CL\cite{zhou-etal-2022-knn}. We provide a more comprehensive comparison and implementation details of these models in the Appendix.

%OpenMax\cite{bendale2016towards}, DeepUnk\cite{lin-xu-2019-deep}, Energy\cite{ouyang-etal-2021-energy},

\begin{table*}[t]
\centering
\resizebox{1.0\textwidth}{!}{%
    \begin{tabular}{c|cc|cc|cc|cc|cc}
    \hline
    \multirow{2}{*}{Models} & \multicolumn{2}{|c}{CLINC-Full} &   \multicolumn{2}{|c}{Banking-25\%} &  
    \multicolumn{2}{|c}{Banking-75\%} &  
    \multicolumn{2}{|c}{Stackoverflow-25\%} &  
    \multicolumn{2}{|c}{Stackoverflow-75\%}\\\cline{2-11}
        ~ & IND F1 & OOD F1 & IND F1 & OOD F1 & IND F1 & OOD F1 & IND F1 & OOD F1 & IND F1 & OOD F1\\ \hline
        SCL\cite{zeng-etal-2021-modeling}  &90.03 &68.21 & 63.32 & 75.82 & 86.56 & 67.51 & 82.45 & 94.09 & 85.76 & 57.46 \\ \hline
        Energy\cite{ouyang-etal-2021-energy}  & 91.23  & 75.93 & - & - & - & - & - & - & - & - \\ \hline
        ADB\cite{Zhang2021DeepOI}  &90.94  &76.52 & - & - & - & - & - & - & - & -  \\ \hline
        KNN-CL\cite{zhou-etal-2022-knn} &92.61 &76.36 & 76.44 & 90.19 & 87.41 & 67.66 & 79.39 & 92.70 & 87.92 & 74.20 \\ \hline
        UniNL(Ours)  &\textbf{93.58}	&\textbf{78.92} & \textbf{78.03} & \textbf{92.46} & 87.13 & \textbf{69.44} & \textbf{87.70} & \textbf{96.33} & \textbf{88.19} & \textbf{74.54} \\ \hline
    \end{tabular}%
}
%\vspace{-0.25cm}
\caption{The performance of our UniNL compared with previous state-of-the-art baselines using BERT.}
%\vspace{-0.65cm}
\label{sota}
\end{table*}

\subsection{Main Results}
Table \ref{main} show the main results on BiLSTM. Our proposed UniNL significantly outperforms all the baselines, which shows that aligning IND pre-training objectives with OOD scoring functions helps improve OOD detection. For example, for OOD metrics, our UniNL outperforms the previous state-of-the-art method SCL+GDA \cite{zeng-etal-2021-modeling} by 5.28\%(Recall) and 8.85\%(F1) on CLINC-Full, 4.54\%(Recall) and 8.96\%(F1) on CLINC-Small. Compared to CE and SCL, KNCL shows significant improvements under all the scoring functions. And our proposed KNN-based scoring function also outperforms previous methods MSP, LOF and GDA. For IND metrics, we find there is no significant difference, which denotes UniNL improves OOD performance without hurting IND classification. Table \ref{sota} also proves our UniNL achieves the state-of-the-art with the same BERT backbone as baselines. All the results show the effectiveness of our proposed KNCL pre-training loss and KNN-based scoring function. Learning unified neighborhood knowledge is beneficial to OOD detection.

% Compared with SCL and CE, we use KNCL objectives for IND pre-training, which shows significant improvement under different detection methods. Besides, our proposed KNN-based detection method consistently outperforms all previous detection methods MSP, LOF and GDA. It shows that the neighborhood discriminative feature is beneficial to distinguish IND and OOD.

%\vspace{-0.2cm}
\subsection{Qualitative Analysis}
\label{Qualitative}

\begin{figure}[t]
    \centering
    \resizebox{.3\textwidth}{!}{
    \includegraphics{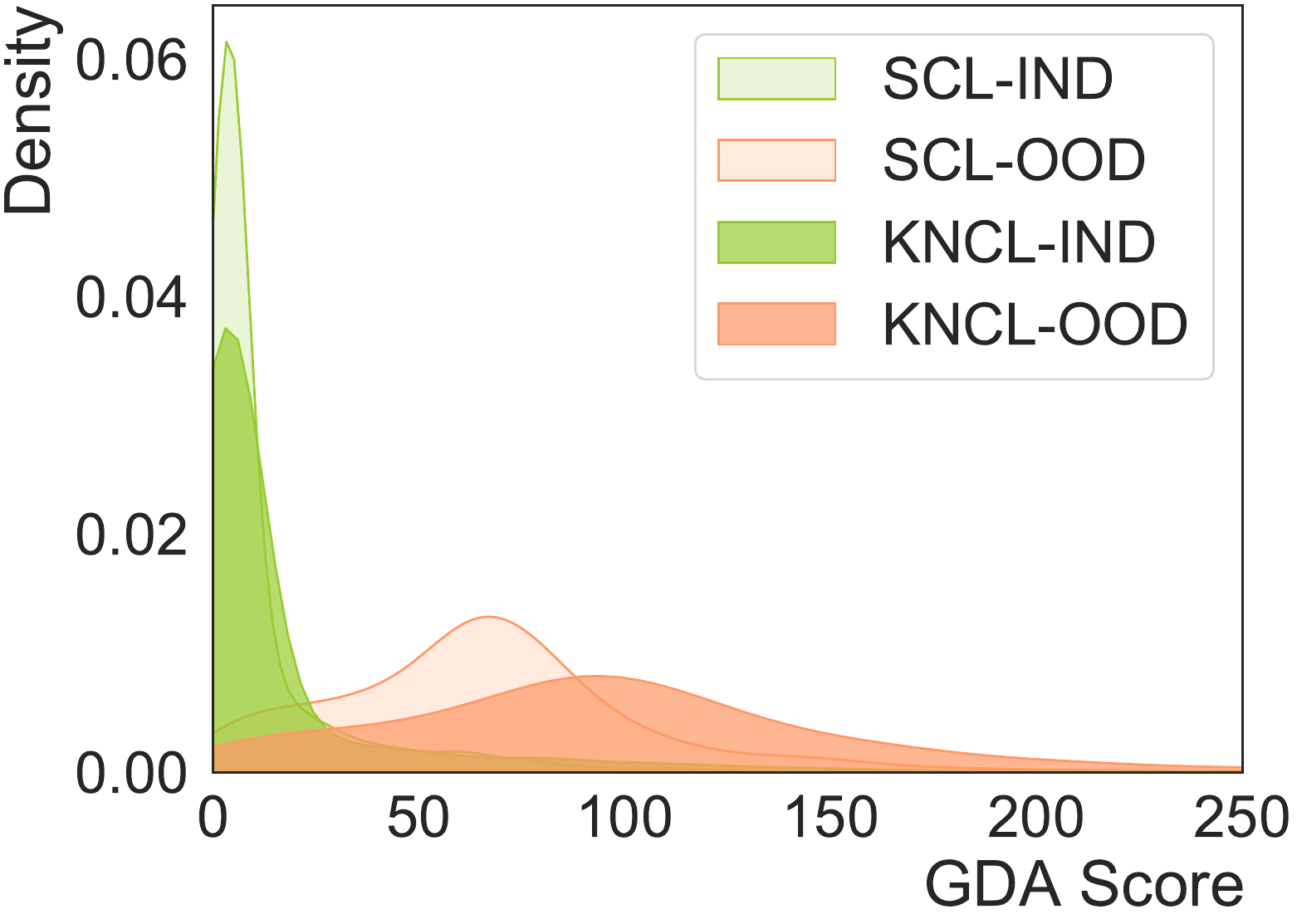}}
    %\vspace{-0.3cm}
    \caption{GDA score distribution curves of IND and OOD data using different IND pre-training losses SCL and KNCL.}
    \label{fig:gda_score}
    %\vspace{-0.5cm}
\end{figure}

\begin{figure}[t]
    \centering
    \resizebox{.45\textwidth}{!}{
    \includegraphics{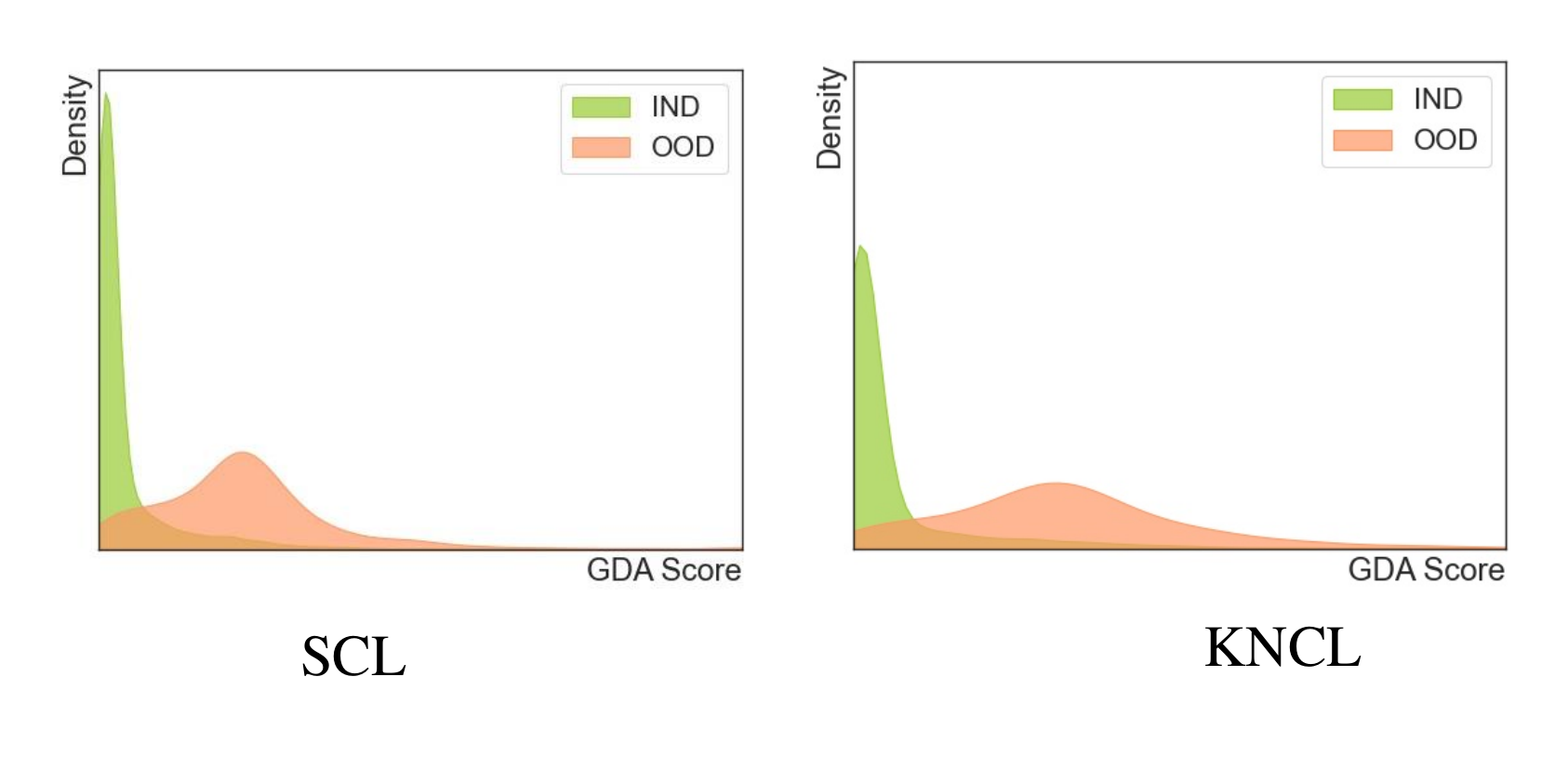}}
    %\vspace{-0.7cm}
    \caption{Confidence score distribution curves of IND and OOD data using different scoring functions GDA and KNN.}
    \label{fig:detection_score}
    % \vspace{-0.6cm}
\end{figure}

\begin{figure}[t]
    \centering
    \resizebox{.35\textwidth}{!}{
    \includegraphics{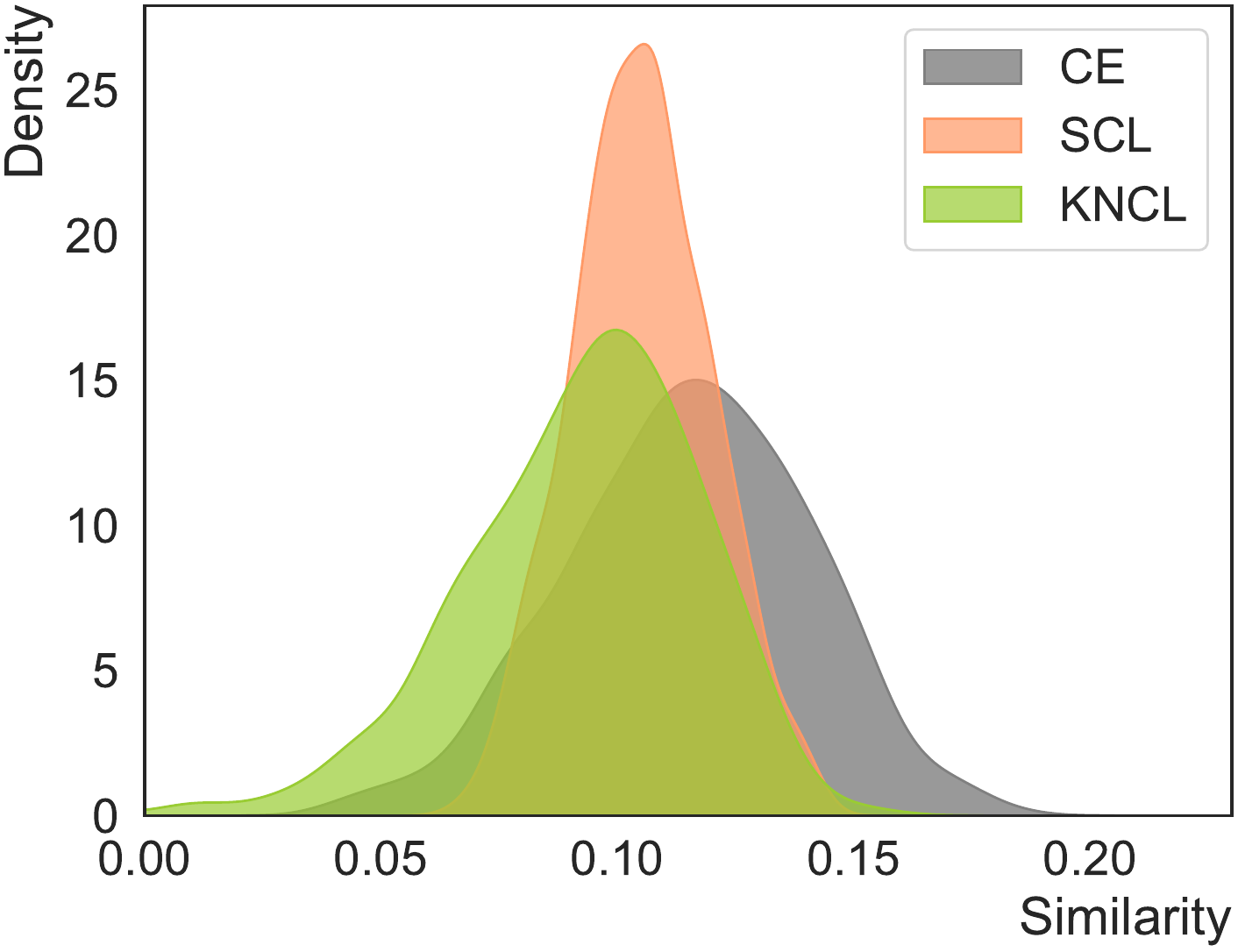}}
    %\vspace{-0.3cm}
    \caption{Similarity score distribution between OOD and IND (The smaller the similarity, the easier it is to distinguish between IND and OOD).}
    \label{fig:similarity}
    \vspace{-0.4cm}
\end{figure}

\textbf{Effect of KNCL} To show the effect of KNCL, we adopt GDA as the score function and compare the GDA score distribution curves of IND and OOD data under different pre-training objectives in Fig \ref{fig:gda_score}. The smaller overlapping area of IND and OOD curves means better performance. We find that KNCL makes the aliasing area of IND and OOD smaller, and improves OOD F1 by 2.69\% compared to SCL. This proves that neighborhood discriminative features help distinguish IND from OOD.

%\subsection{Effect of KNN-based detection}
\textbf{Alignment between KNCL and KNN} With KNCL as the pre-training objective, we compare the GDA and KNN score distribution curves for IND and OOD data, as shown in Fig \ref{fig:detection_score}. The more separated the IND and OOD distribution curves are, the more favorable it is for OOD  detection. It can be seen that the KNN scores have better effect on distinguishing IND and OOD, which also indicates that aligning the IND pre-training objective and the OOD scoring function helps to improve the performance of OOD detection.

\textbf{Why Alignment Works Well} To discuss why the KNCL representation learning objective matches the KNN-based scoring function, we compare the cosine similarity distance between OOD and IND under different representation learning objectives in Fig \ref{fig:similarity}. For each OOD sample in the test set, we calculate the average of its cosine similarity scores with the K-nearest neighbor IND samples from training data, and obtain the cosine similarity score distribution curve. We find that KNCL can decrease the average similarity between OOD and IND, which has a boosting effect on the KNN-based scoring function.

%Using CE, SCL, KNCL for representation learning, the average similarity scores from OOD to IND are 0.11, 0.10, 0.09, respectively.

\begin{figure*}[t]
    \centering
    \centering
    \subfigure[SCL]{
        \includegraphics[scale=0.45]{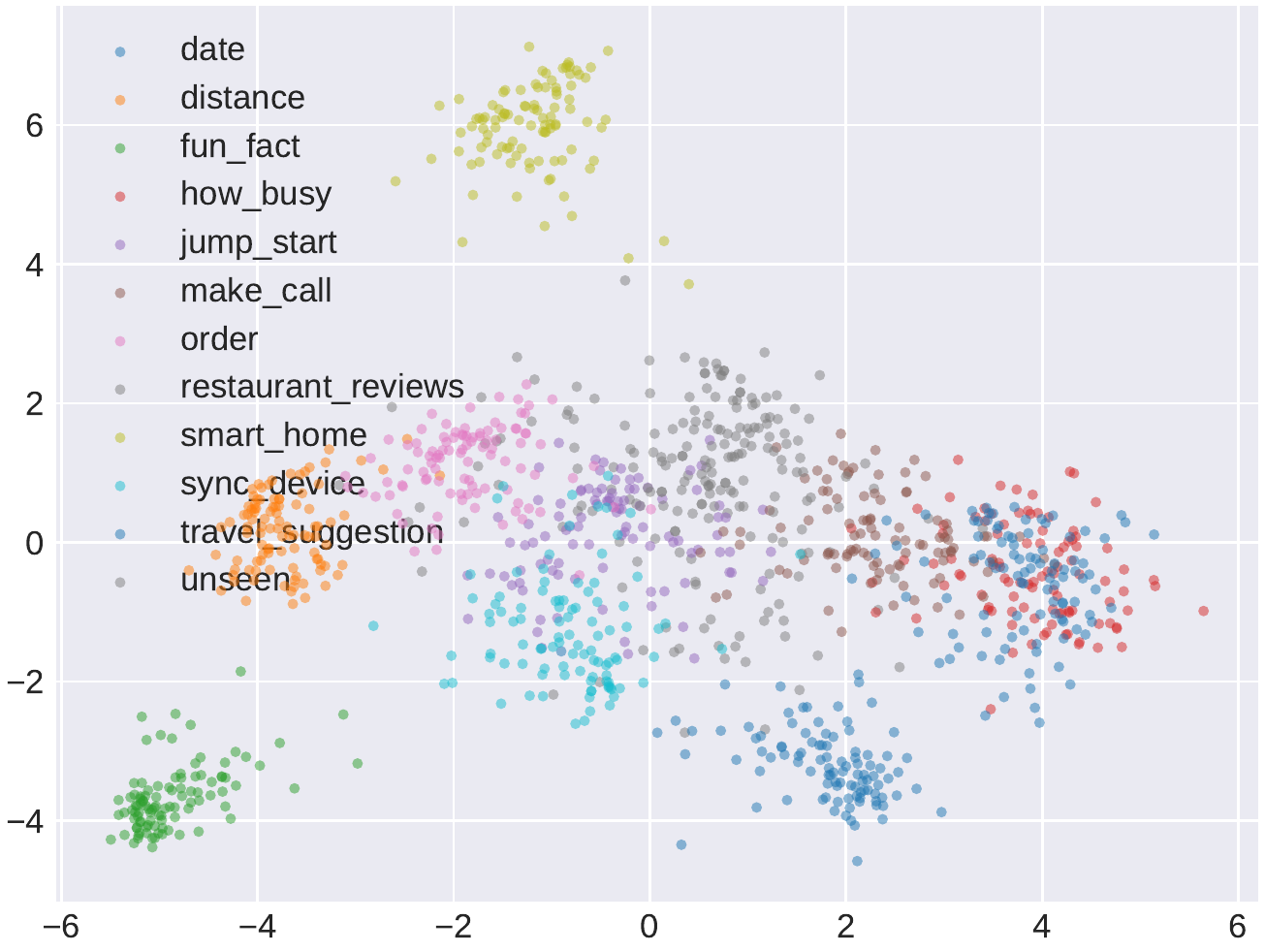}
    }
    \subfigure[KNCL]{
        \includegraphics[scale=0.45]{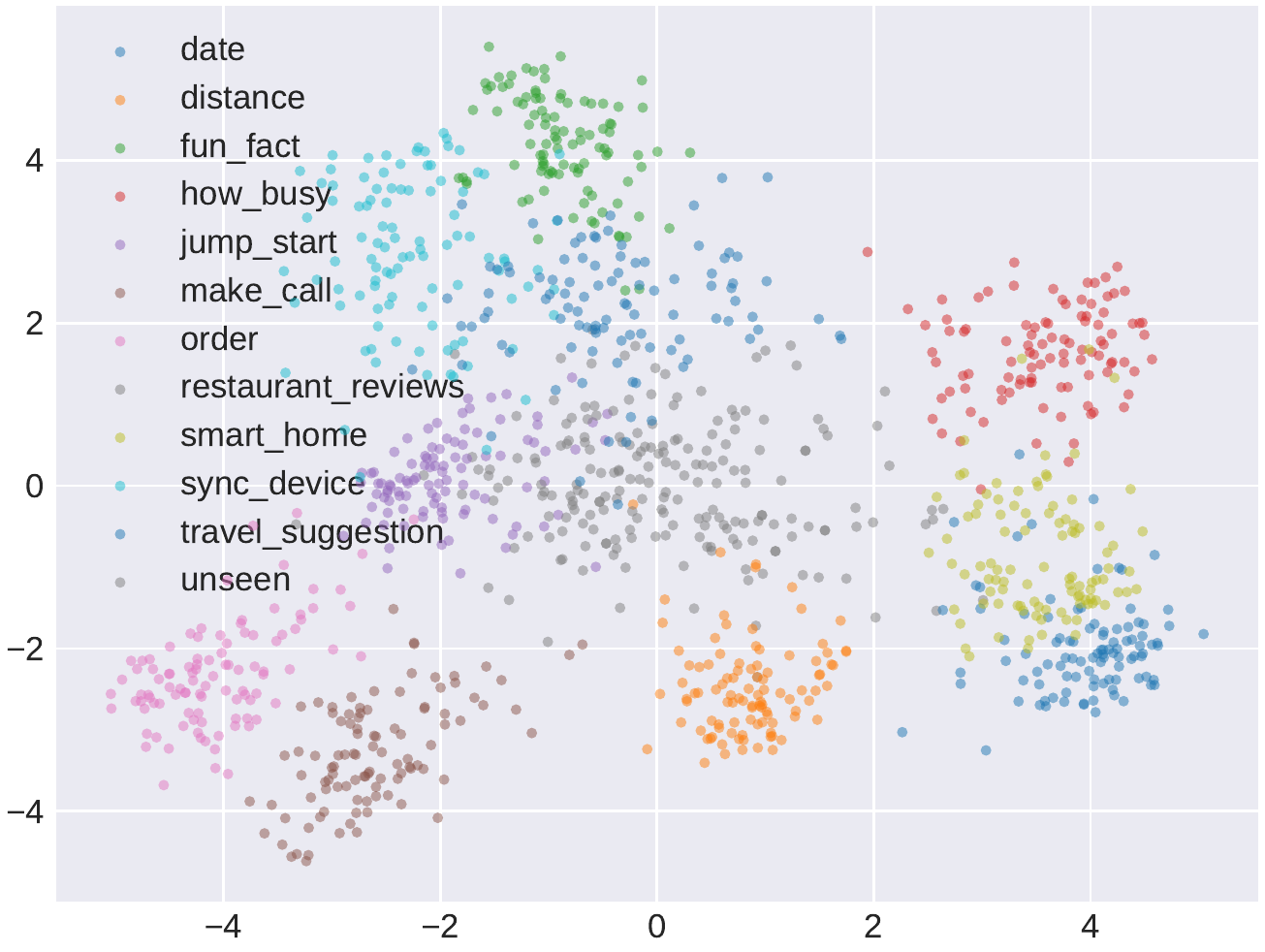}
    }
    %\vspace{-0.45cm}
    \caption{Visualization of IND and OOD intents (The "unseen" legend label means OOD intents).}
    \label{visualization}
    %\vspace{-0.8cm}
\end{figure*}

\begin{figure} [t]
    \centering
    \includegraphics[scale=0.35]{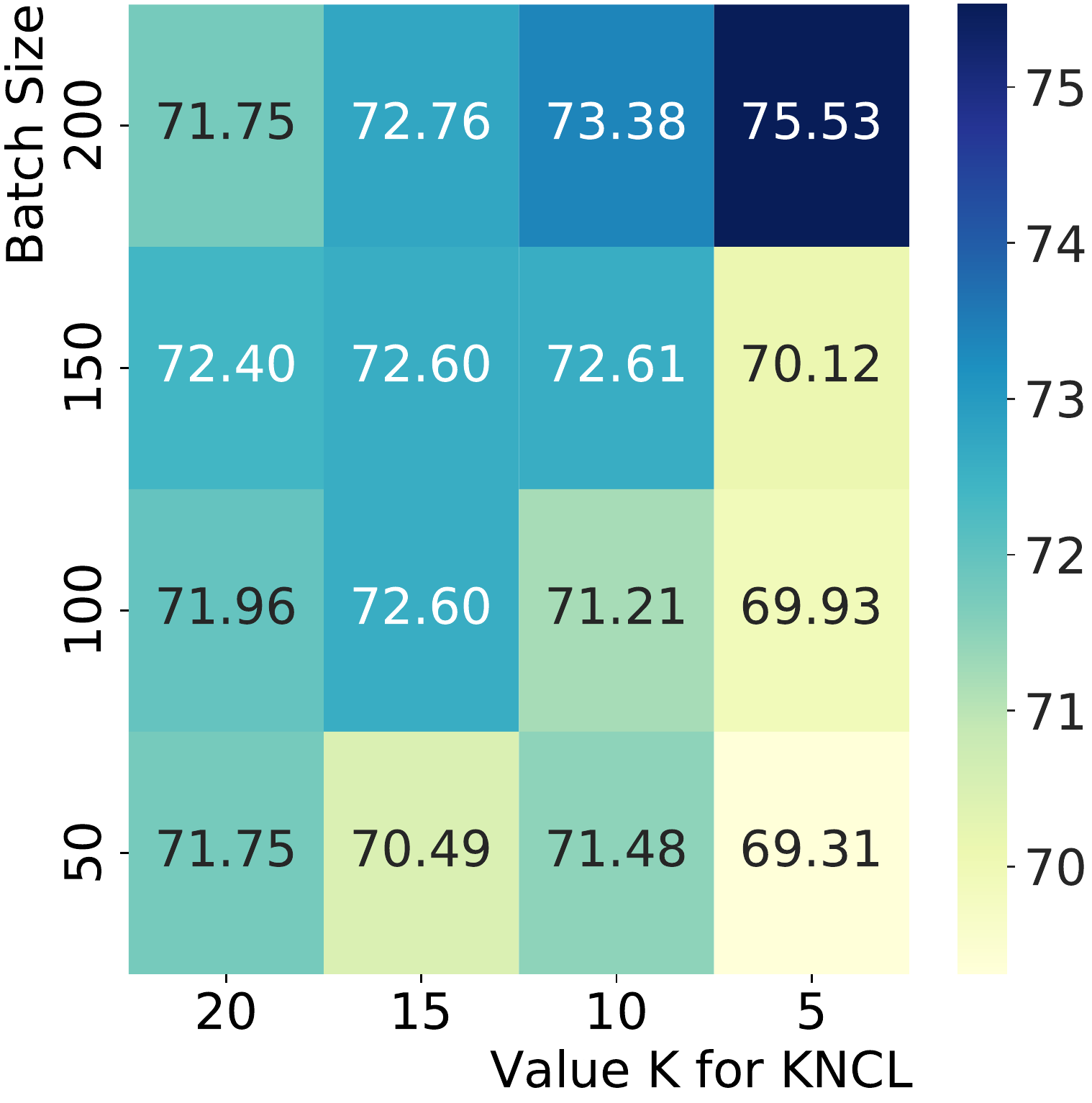}
    \caption{Effect of batch size and K value of KNCL on clustering performance. The larger the number is, the deeper the color is.}
    \vspace{-0.2cm}
    \label{fig:batch_k}
\end{figure}

\begin{figure} [t]
    \centering
    \includegraphics[scale=0.4]{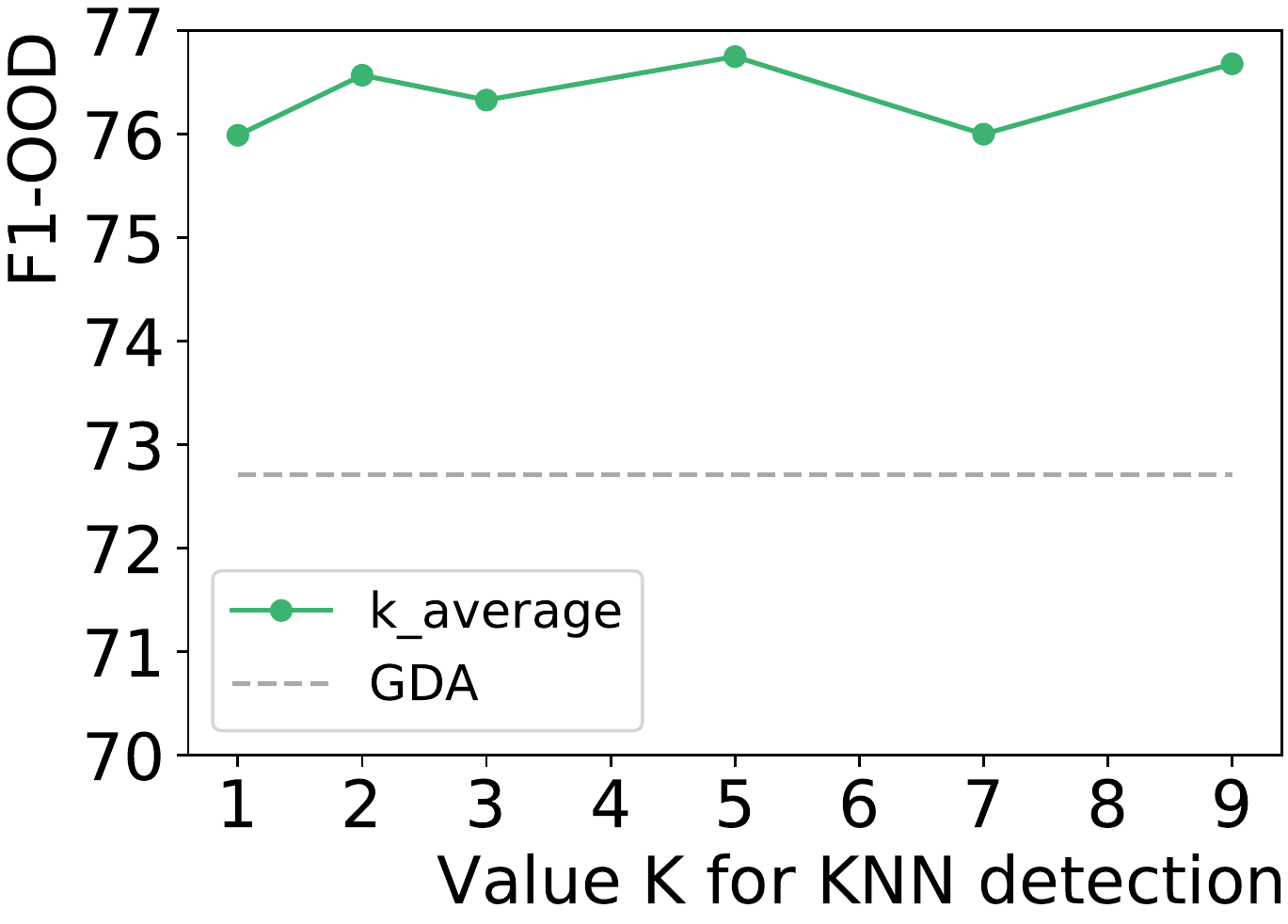}
    \caption{Effect of  K value of KNN-based score function on clustering performance.}
    \label{fig:k_KNN}
\end{figure}

\textbf{The effect of the K value for KNCL} The KNCL pre-training objective requires a reasonable choice of K value. We compare the OOD F1 scores under different batch sizes and K values, as shown in Fig \ref{fig:batch_k}. We found that the batch size will affect the choice of K value. When the batch size is larger, a smaller K value can achieve better results; when the batch size is smaller, a larger K value is required to achieve good performance. We argue that this is because the larger the batch size is, the better it can approximate the real distribution of the entire dataset, and a smaller K value can simulate the real neighborhood distribution.

\textbf{The effect of the K value for KNN score} The K value of the KNN-based score function is also an important hyperparameter, and we compare the effect of different K values on the OOD detection performance, as shown in Fig \ref{fig:k_KNN}. It can be seen that this K value has little effect on the OOD detection performance, which illustrates the robustness of our proposed KNN-based detection method.

%Specifically, given each OOD sample in the test set, we first find its K nearest neighbor IND sample set from training data according to the cosine similarity of the embedding, and then average similarity scores with the KNN set as similarity measure between the OOD and IND. The results are shown in Fig \ref{fig:similarity}. We compared the similarity score distribution curves of different IND pre-training objectives CE, SCL and KNCL. The closer the similarity distribution curve is to the left, the smaller the overall similarity between OOD and IND, that is, the greater the distance between OOD and IND. It can be seen that KNCL can make the distribution curve closer to the left, which also shows that KNCL is more helpful to distinguish IND and OOD. In addition, KNCL increases the distance between OOD and IND, which has a boosting effect on KNN-based detection methods.

\textbf{Visualization} Fig \ref{visualization} displays IND and OOD intent visualization for different IND pre-training objectives SCL and KNCL. It shows that the SCL objective will confuse the OOD samples with the "jump\_start" and "make\_call" intent types, while KNCL can distinguish them well. This proves that KNCL can better distinguish IND and OOD by modeling neighborhood discriminative features, which is beneficial to improving the performance of the KNN-based scoring function.

\section{Conclusion}
In this paper, we focus on how to align representation learning with the scoring function to improve OOD detection performance. We propose a unified neighborhood learning framework (UniNL) to detect OOD intents, in which a KNCL objective is employed for IND pre-training and a KNN-based score function is used for OOD detection. Experiments and analyses confirm the effectiveness of UniNL for OOD detection. We hope to provide new guidance for future OOD detection work.

%in which a KNCL objective is employed for IND pre-training and a KNN-based score function is used for OOD detection.

\section*{Limitation}
This paper mainly focuses on how to align the representation learning and scoring functions to achieve better OOD detection performance. Thus we follow similar experiment settings as previous work. 
However, similar to these works, we only experiment with datasets in the field of intent recognition. Actually, OOD detection has applications in a wider range of NLP topics, such as relation classification, entity recognition, text classification, etc. We will try our proposed method on more NLP topics in the future to verify the universality.

\section*{Acknowledgements}
We thank all anonymous reviewers for their helpful comments and suggestions. This work was partially supported by National Key R\&D Program of China No. 2019YFF0303300 and Subject II No. 2019YFF0303302, DOCOMO Beijing Communications Laboratories Co., Ltd, MoE-CMCC "Artifical Intelligence" Project No. MCM20190701.

\bibliography{anthology,custom}
\bibliographystyle{acl_natbib}

\appendix

\section{Experiment Setups}
\label{setups}

\subsection{Datasets}
We perform experiments on four public benchmark OOD datasets, CLINC-Full, CLINC-Small \cite{Larson2019AnED}, Banking \cite{casanueva2020efficient} and Stackoverflow \cite{xu2015short}. We show the detailed statistics of the datasets in Table \ref{dataset} .

\begin{table}[h]
\centering
\resizebox{0.48\textwidth}{!}{%
\begin{tabular}{l|cccc}
\hline
Statistic & CLINC-Full & CLINC-Small & Banking & Stackoverflow \\ \hline
Avg utterance length & 9  & 9 & 12 & 10           \\
Intents              & 150        & 150  & 77 & 20       \\
Training set size    & 15100      & 7600  & 9003 & 12000      \\
Training samples per class  & 100      & 50 & - & - \\
Training OOD samples amount & 100      & 100 & - & - \\
Development set size & 3100       & 3100 & 1000 & 2000        \\
Development samples per class & 20       & 20 & - & -      \\
Development OOD samples amount & 100       & 100 & - & - \\
Testing Set Size     & 5500       & 5500  & 3080 & 6000      \\
Testing samples per class & 30       & 30  & - & -      \\
Development OOD samples amount & 1000       & 1000 & - & - \\ \hline
\end{tabular}
}
\caption{Statistics of the CLINC datasets.}
\vspace{-0.5cm}
\label{dataset}
\end{table}

\subsection{Baselines}
We compare UniNL with different pre-training objectives and different scoring functions. For the feature extractor, we use the same BiLSTM\cite{Hochreiter1997LongSM} or BERT \cite{Devlin2019BERTPO} as backbone. We compare our training objective KNCL with CE and SCL\cite{zeng-etal-2021-modeling}, and scoring function KNN with MSP\cite{Hendrycks2017ABF}, LOF\cite{Lin2019DeepUI} and GDA\cite{xu-etal-2020-deep}. Besides, we also compare our model with the following state-of-the-art baselines, Energy\cite{ouyang-etal-2021-energy} and ADB\cite{Zhang2021DeepOI}. We supplement the relevant baseline details as follows:

\noindent\textbf{MSP} (Maximum Softmax Probability)\cite{Hendrycks2017ABF} uses maximum softmax probability as the confidence score. If the score is lower than a fixed threshold, the query is regarded as OOD. In this paper, we use the best IND macro F1 scores on the validation set to calculate the threshold adaptively.

\noindent\textbf{LOF} (Local Outlier Factor)\cite{Lin2019DeepUI} uses the local outlier factor to detect unknown intents. It detects OOD by comparing the local density of a test query with its k-nearest neighbor’s. If a query’s local density is significantly lower than its k-nearest neighbor’s, it is more likely to be regarded as OOD.

\noindent\textbf{GDA} (Gaussian Discriminant Analysis)\cite{xu-etal-2020-deep} is a generative distance-based classifier to detect OOD samples. It estimates the class-conditional distribution on feature spaces of DNNs via Gaussian discriminant analysis, and then applies Mahalanobis distance to measure the confidence score. When estimating the class conditional distribution with labeled IND data, we assume that it follows a Gaussian distribution. However, the representation space modeled by the cross-entropy objective cannot actually satisfy the Gaussian distribution assumption.

%To avoid over-confidence problems, they combine the advantages of DNNs and Gaussian discriminant analysis and apply Mahalanobis distance to measure the confidence score. 

\noindent\textbf{SCL} \cite{zeng-etal-2021-modeling} uses a supervised contrastive learning objective to minimize intra-class variance by pulling together in-domain intents belonging to the same class and maximize inter-class variance by pushing apart samples from different classes. To keep fair comparison, we follow the original paper using GDA as score function.

\noindent\textbf{Energy} \cite{ouyang-etal-2021-energy} maps a sample $x$ to a single scalar called the energy. It uses the threshold on the energy score to consider whether a test query belongs to OOD.

\noindent\textbf{ADB} \cite{Zhang2021DeepOI} learns adaptive decision boundary using a loss function to balance both the empirical risk and the open space risk.

\noindent\textbf{KNN-CL} \cite{zhou-etal-2022-knn} is concurrent work with our UniNL. It proposes a KNN-based contrastive loss for IND pre-training, which is conceptually similar to our KNCL. But our implementations are different: KNN-CL selects k-nearest neighbors from samples of the same class as positives and uses samples of the different classes as negatives. Our KNCL only uses the k-nearest neighbors of an anchor as the positive and negative set, which is more efficient and doesn't require a large momentum queue as KNN-CL. Moreover, we aim to align IND pre-training representation objectives with OOD scoring functions instead of proposing a better IND pre-training loss.

\subsection{Implementation Details}
To conduct a fair comparison, we follow a similar evaluation setting as \citet{zeng-etal-2021-modeling}. We use the public pre-trained GloVe 300 dimensions embeddings\cite{pennington2014glove} and BERT-uncased model to embed tokens. We set the learning rate to 1e-03 for LSTM and 1e-04 for BERT\cite{Devlin2019BERTPO}. We use Adam optimizer\cite{kingma2014adam} to train our model and set the dropout rate to 0.5. In the training stage, we firstly conduct 100 epochs of K-nearest neighbor contrastive training, and then 10 epochs with CE. The K value of KNCL objective is set to 5. When performing KNCL, we first select the KNN set on the original view, and then extend the KNN set to its augmented view to participate in the calculation of the contrast loss.
In the test stage, we set the K value of KNN scoring function to 5. And we use the best IND macro F1 scores on the validation set to calculate the threshold adaptively. To avoid randomness, we average results over 5 random runs.
%Each result of the experiments is tested 5 times under the same setting and gets the average value. 
Table \ref{tab:implementation_detail} shows the comparison of the computational efficiency for different pre-training objectives and scoring functions. For training efficiency, we report the running time of each epoch; for inference efficiency, we report the total running time on the entire test set of CLINC-Full.
It can be seen that the inference efficiency of our proposed KNN-based scoring function has been greatly improved. Compared with GDA, the efficiency of KNN score is increased by 14.13 times. The training efficiency for KNCL objective function has a slight decrease due to the need for K-nearest neighbor search, but it gives about 2\%-4\% OOD F1 performance improvement for all scoring functions.
%The training time for CE is about 6 s/epoch using Glove+LSTM, 10 s/epoch for SCL and 20 s/epoch for KNCL. During the OOD detection stage, the computation of our KNN scores lasts for 0.46s, which greatly improved the 6.50s of GDA and 104.60s of LOF. 
All experiments use a single Tesla T4 GPU(16 GB of memory). 

\begin{table}[t]
\centering
\resizebox{0.45\textwidth}{!}{
    \begin{tabular}{c|c|c|c}
    \hline
        \multicolumn{2}{c|}{Method}  & \makecell[c]{training time \\ (seconds / epoch)} & \makecell[c]{inference time \\ (seconds)} \\ \hline
        \multirow{3}{*}{Training objectives} & CE & 6  & - \\ 
        & SCL  & 10   & -  \\ 
        & KNCL(ours)  & 20   & - \\ \hline
        \multirow{3}{*}{Score functions} & LOF & - & 104.60  \\ 
        & GDA  & - & 6.50   \\ 
        & KNN(ours)  & - & 0.46   \\ \hline
    \end{tabular}}
%\vspace{-0.2cm}
\caption{The comparison of the computational efficiency for different pre-training objectives and scoring functions.}
\vspace{-0.2cm}
\label{tab:implementation_detail}
\end{table}

% TO DO: (1) 用个表展示效率; (2) 补一个消融

\begin{table}[t]
\centering
\resizebox{0.48\textwidth}{!}{%
    \begin{tabular}{l|cccc}
    \hline
    \multirow{2}{*}{Training} &   \multicolumn{4}{|c}{CLINC-Full} \\\cline{2-5}
        ~ & IND ACC & IND F1 & OOD Recall & OOD F1\\ \hline
        only KNCL  & - & - &62.41 &60.31 \\ \hline
        only CE &90.30 &\textbf{88.89} &67.03 &73.32 \\ \hline
        CE+KNCL  &85.30 &83.12  &71.55 &72.57 \\ \hline
        multitask  &90.38  &87.47  &69.25  &73.16  \\ \hline
        KNCL+CE  &\textbf{91.24}  &88.28	&\textbf{72.08}	&\textbf{76.53} \\ \hline
    \end{tabular}%
}
%\vspace{-0.25cm}
\caption{Comparison of different training strategies in the IND pre-training stage.}
\vspace{-0.65cm}
\label{different_ways}
\end{table}

\section{Ablation Study}
\label{ablation}
% todo: 补充下ce kncl的融合方式的差异

In order to verify which training strategy is the most effective in the IND pre-training stage, we compared the combination of different training objectives, and the results are shown in Table \ref{different_ways}. We conduct experiments on the CLINC-Full dataset, using BiLSTM as encoder and KNN as scoring function. Only CE is the baseline that only uses the cross-entropy loss function to train the feature extractor. Only KNCL means that we use KNCL objective for representation learning. KNCL+CE means that we first train the feature extractor using KNCL, and then fine-fune it using CE loss. CE+KNCL means that we first train the network by minimizing cross-entropy loss, and then conduct K-nearest neighbor contrastive learning. Besides, we also compare the simple multitask paradigm, which simply adds the CE and KNCL objective functions for joint optimization. The results show that the best performance can be achieved by first learning the neighborhood discriminative knowledge with KNCL and then fine-tuning the model with CE \footnote{When only KNCL is used to train the feature extractor, softmax classifier cannot be used for IND classification, so we do not report relevant IND results here.}.

\end{document}